\documentclass{article}

\usepackage[final]{tccml_neurips_2020}

\usepackage[utf8]{inputenc} 
\usepackage[T1]{fontenc}    
\usepackage{hyperref}       
\usepackage{url}            
\usepackage{booktabs}       
\usepackage{amsfonts}       
\usepackage{nicefrac}       
\usepackage{microtype}      
\usepackage{graphicx}
\usepackage{xfrac}
\usepackage[allow-number-unit-breaks]{siunitx}

\title{Residue Density Segmentation for Monitoring and Optimizing Tillage Practices}

\author{%
  Jennifer Hobbs\thanks{corresponding author} \\
  Intelinair, Inc.\\
  Champaign, IL 61820 \\
  \texttt{jennifer@intelinair.com} \\
  
  \And
  Ivan Dozier\\
  Intelinair, Inc.\\
  Champaign, IL 61820 \\
  \texttt{ivan@intelinair.com} \\
  
   \And
  Naira Hovakimyan\\
  University of Illinois at Urbana-Champaign\\
  Intelinair, Inc.\\
  Champaign, IL 61820\\
  \texttt{naira@intelinair.com}
  \\[-3.0ex]
}

\begin{document}

\maketitle

\begin{abstract}
  ``No-till'' and cover cropping are often identified as the leading simple, best management practices for carbon sequestration in agriculture.
  However, the root of the problem is more complex, with the potential benefits of these approaches depending on numerous factors including a field's soil type(s), topography, and management history.
  Instead of using computer vision approaches to simply classify a field as till vs. no-till, we instead seek to identify the degree of residue coverage across a field through a probabilistic deep learning segmentation approach to enable more accurate analysis of carbon holding potential and realization. This approach will not only provide more precise insights into currently implemented practices, but also enable a more accurate identification process of fields with the greatest potential for adopting new practices to significantly impact carbon sequestration in agriculture. 
\end{abstract}

\section{Residues, Tillage Practices, and Nutrient Management}
Carbon sequestration is one of the primary topics raised in discussions around agriculture and climate change.  
Soils have the capacity to be enormous carbon sources or sinks with farm management practices significantly impacting how much carbon is held in the soil~\cite{houghton2003estimates}.
Past agricultural management decisions in the US have depleted global soil organic carbon (SOC) by as much as $72.8\pm13.2$ billion US tn~\cite{lal1999soil}.
US cropland which covers roughly $157\frac{M}{ha}$ is estimated to have the capacity to sequester $0.3-0.5\frac{Mg}{ha}$ of carbon/year for a total potential of $45-98 Tg$carbon/year~\cite{chambers2016soil, lal2015sequestering}.
Altering past practices to bring carbon out of the atmosphere and into the soil helps to both mitigate greenhouse gasses, reduce negative agricultural contributions to the environmental, and improves soil health and water holding capacity~\cite{smith2012soils, chambers2016soil, wade2015conservation}.

Many initiatives around carbon sequestration for cropland are heavily focused around tillage practices~\cite{cole1997global}.
Residues consist of crop biomass such as dried leaves and stalks leftover from harvest; these residues contain key nutrients that the plants had absorbed during the season.
By reincorporating these residues back into the soil, usually via tilling, farmers are able to recycle those nutrients: as residues decompose, nutrients re-enter the soil, fueling the next year's crops.
In contrast, ``no-till'' and alternative tillage practices limit the amount of tillage conducted.  
Maintaining surface residues has numerous benefits including increasing SOC and water capacity, increasing porosity, preventing erosion, and enhancing soil stability, especially when used in combination with cover crops~\cite{wade2015conservation}.
Switching to no-till additionally requires one-less step in the farming life-cycle, saving labor time as well as reducing fuel usage.
However, residues can cause keep nutrients tied up in unusable forms, harbor pest and diseases, and inhibit emergence, leading to significant loss of yield if not adequately managed~\cite{gollany2004nitrogen}.

As a result, adoption of no-till and reduced-tillage practices vary widely across regions and crops, with only $\sim20\%$ of farmland using no-till practices continuously~\cite{wade2015conservation, creech2017saving}.
While many associate no-till and cover cropping as the key, beneficial approaches in carbon sequestration and erosion prevision, the impact of various tillage practices is far more complicated; the amount of carbon which can be sequestered with these practices can vary widely based on soil composition, moisture-levels, topography, and other management decisions~\cite{swan2018comet, chambers2016soil}.
The economic benefit of these practices must be established in an accurate, personalized manner for each farm in order to promote widespread trust and adoption.

\begin{figure}
    \centering
    \includegraphics[width=.8\linewidth, trim={0 0cm 0 0cm}]{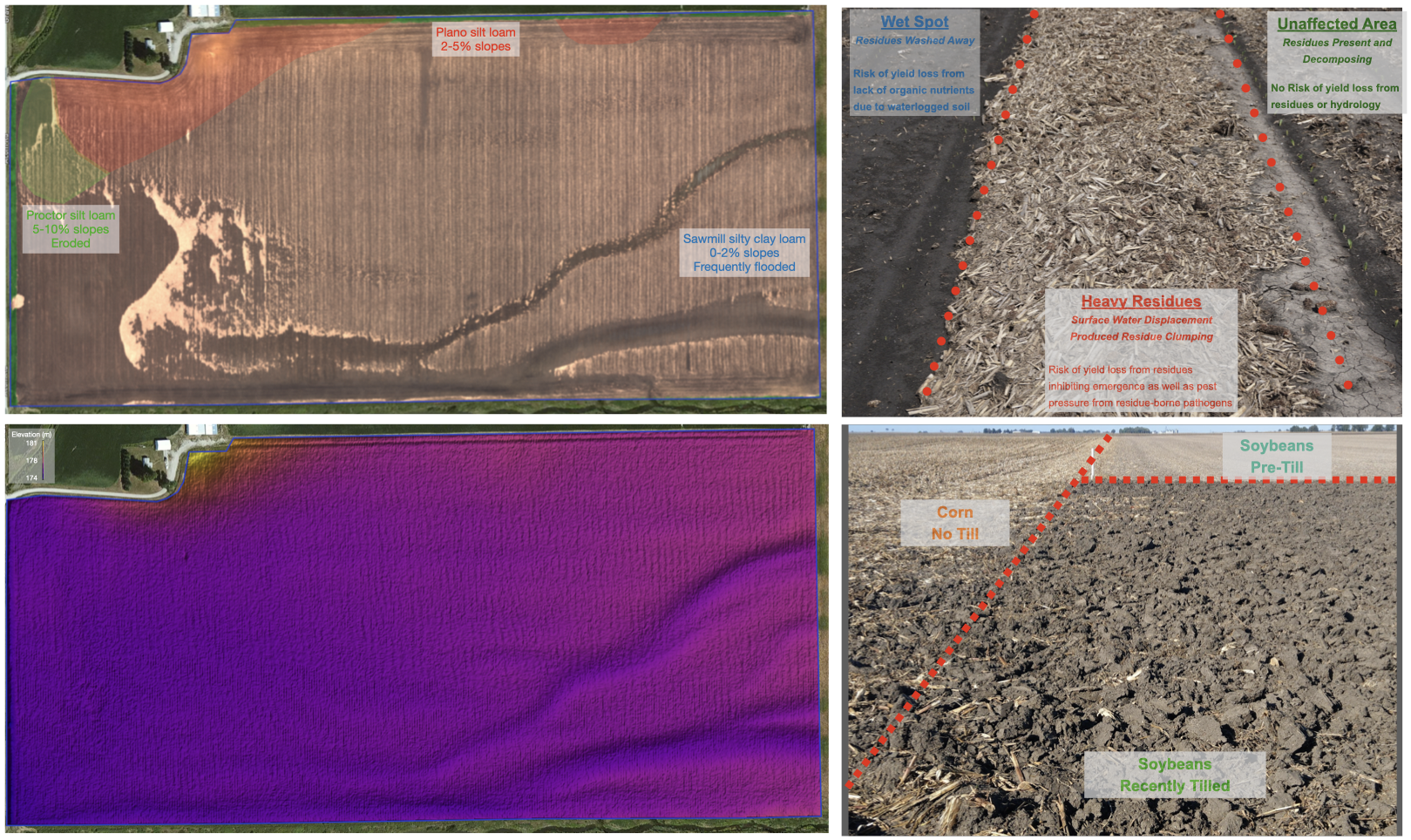}
    \caption{(Top Left) Aerial image of a field with highly variable residue density. Such variability is due to a combination of management practices, soil composition, and topography (Bottom Left).  At ground-level (Right), we see these same factors as well as crop-type and time-since-tillage impact the field's appearance. }
\label{fig:aerial}
\end{figure}

The visual impact of these management choices manifest themselves in complex ways across the field as seen in Figure~\ref{fig:aerial}.
Capturing and understanding all of these contributing factors is critical for accurately assessing the impact of tillage practices on a particular farm as well as encouraging broad adoption of these management practices; a simple classification of till/no-till is not enough. 
Therefore we segment the field into different levels of residue coverage to provide a fine-grained map of this biomass layer.
This information can be combined with soil, hydrological, and other models to more accurately determine the opportunity for carbon sequestration on a given farm parcel and the impact of tillage practices to capitalize on those storage capacities. 
Identifying the density of residue further helps farmers more effectively manage residues across their field and enables novel precision tillage practices.
As opposed to treating tillage as a binary management practice, farmers can take a targeted approach to till specific areas of their fields in the way that best addresses carbon sequestration, yield, nutrient, and erosion risks: casting these tillage practices not as a choice between economic and environmental needs, but as a strategic plan which maximizes both.   


\section{Approach}
\subsection{Data Acquisition and Annotation}
The appearance of residues in a field is strongly dictated by the crop type.
Identification of the crop planted in the previous season can be accomplished from publicly available low-resolution ($>$10m/pixel) satellite imagery which provides flexibility in allowing crop-identification to be performed months after the crop has been harvested from the field; knowing which crops to analyze in the next season does not need to be known far in advance.
Crop identification from satellite is a widely studied problem and can be solved with any number of computer vision and deep learning approaches~\cite{kussul2017deep_ct, zhong2019deep_ct}.

While identification of till vs. no-till \textit{may} be feasible from low-resolution satellite imagery, segmentation of residue levels is not.  
Therefore we will collect high-resolution ($<$1m/pixel) RGB + NIR aerial imagery over the fields of interest.
During initial research phases this can be done with drones and planes, but can be extended to other collection methods (e.g. high-resolution satellite), to scale the method more broadly in the long-term.

\begin{figure}
    \centering
    \includegraphics[width=.8\linewidth, trim={0 0cm 0 0cm}]{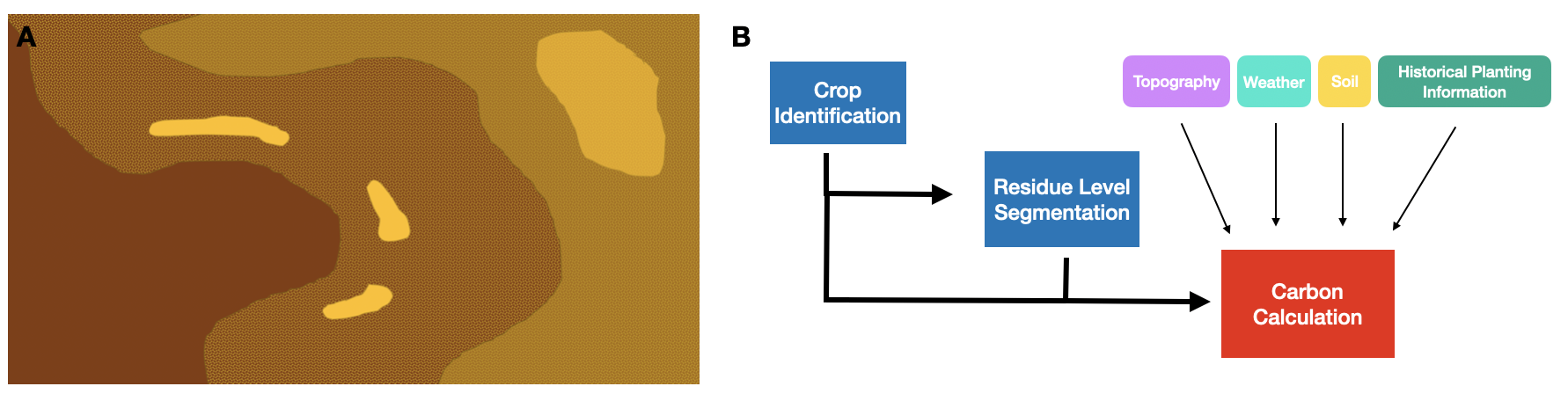}
    \caption{(A) Sample ``image'' showing the five different levels of annotations to be captured from no residue (dark brown) through ponding (yellow). (B) Overall flow of how information is captured for the final carbon impact calculation.}
\label{fig:method}
\end{figure}

Next, images are annotated for five different residue density levels: none (background), low ($>50\%$ soil visible), moderate ($50-75\%$ soil visible), heavy ($<25\%$ soil visible), and ponding ($0\%$ soil visible with obvious multi-layer buildup) as seen in Figure~\ref{fig:method}.  
Since this is an inherently challenging annotation task due to the ambiguity over where specific borders end, each image will be annotated by multiple annotators so that the distribution over annotators can be learned~\cite{kohl2018probabilistic}.

\subsection{Modeling}
Fully convolutional encoder-decoder neural networks have proven highly successful in many segmentation tasks, including those in agriculture~\cite{fawakherji2019crop, fuentes2019fig, unet}. 
Following these and the probabilistic U-Net approach above, we learn the distribution over the the plausible level segmentation; a five-channel (RGBN + topography) image serves as input into the model and a five-channel (one per level) image is returned.  
The best approach to fuse topography and imagery will be explored~\cite{sheng2020effective}. 

This mapping of residue levels alone provides value as it can be used to alert the farmer to areas which may experience emergence issues due to excessive residues and ponding.
Furthermore, this residue map would be combined with other sources of information such as soil makeup, weather, topography, etc. and passed to downstream calculations and models to compute both the potential as well as achieved carbon sequestered for the given farmland(Figure~\ref{fig:method})~\cite{swan2018comet, blanco2008no}.

\subsection{Promoting Adoption}
The sustainability and conservation goals that alternative and reduced tillage practices promise are only attainable if widely adopted in significant areas.
As with many conservation initiatives, till vs. no-till is often seen as a binary choice between maximizing economic or environmental outcomes, leading to slow or minimal adoption. 
This approach enables us to provide farmers with intelligence about their farm, enabling them to make the best decisions about \textit{where} and \textit{how much} to till for long term sustainability in regards to both the environment and yield.  
The residue map can be further used to alert farmers to ponding boundaries and other areas which could be susceptible to disease, pests, and emergence suppression. 
Recent years have shown how precision agriculture practices around chemical and water applications have led to both economic and environment advances, and this approach will enable the same for tillage practices.

\section{Conclusion}
Reframing the discussion around carbon sequestration for agriculture, not in the overly simplistic terms of till vs. no-till, but as precision residue management, is crucial for identifying the best tactics for a given farm, accurately quantifying the impact of those decisions, as well as promoting adoption.
High-resolution aerial imagery and deep learning approaches will allow us to accurately determine levels of residue across the field; because annotating densities is a difficult challenge, we will use a probabilistic segmentation approach to learn density levels over annotators.  
The final residue map combined with other data layers such as topography and soil type will enable a more complete understanding of the potential as well as realized carbon sequestration opportunities for that field.

\medskip
\small
\bibliography{refs}
\bibliographystyle{ieee_fullname}

\end{document}